\documentclass{article}

\IfFileExists{neurips_2026.sty}{%
  \usepackage[preprint]{neurips_2026}%
}{%
  \usepackage[margin=1in]{geometry}%
  \newlength{\bibsep}%
  \newcommand{\workshoptitle}[1]{}%
}
\workshoptitle{AI for Drug Discovery}

\usepackage[utf8]{inputenc}
\usepackage[T1]{fontenc}
\usepackage{hyperref}
\IfFileExists{references.bib}{}{%
  \renewcommand{\cite}[1]{[\texttt{##1}]}%
}
\usepackage{url}
\usepackage{booktabs}
\usepackage{amsmath}
\usepackage{amsfonts}
\usepackage{array}
\usepackage{tabularx}
\usepackage{graphicx}
\newcommand{\reviewgraphic}[2][]{%
  \IfFileExists{#2}{\includegraphics[#1]{#2}}{%
    \fbox{\parbox[c][1.15in][c]{0.8\linewidth}{\centering\footnotesize
      Figure file not supplied with the two compared sources:\\[2pt]\texttt{\detokenize{#2}}}}%
  }%
}
\usepackage{microtype}
\usepackage{xcolor}
\IfFileExists{soul.sty}{\usepackage{soul}}{}
\usepackage{float}
\IfFileExists{capt-of.sty}{\usepackage{capt-of}}{}

\newif\iftrackchanges
\trackchangesfalse

\definecolor{DarkYellow}{RGB}{145,110,0}

\definecolor{Burgundy}{RGB}{128,0,48}

\title{Beyond Random Splits: Evaluating Drug-Target Affinity Models Under Chemically and Biologically Motivated Distribution Shifts}

\author{
\parbox{\textwidth}{\centering
Minjae Chung, Clara Li, Malar Paavai Muthukumaran,\\
Shaunna Wang, Aniket Ramkrishnan Iyer, Shaun Qien Yeau Tan,\\
Harinishree Sathu, Micky C. Nnamdi, J. Ben Tamo,\\
Benoit Louis Marteau, May Dongmei Wang\\[6pt]
\textit{Georgia Institute of Technology}\\
\texttt{mchung98@gatech.edu}
}
}
\RequirePackage[normalem]{ulem} 
\RequirePackage{color}\definecolor{DiffPurple}{RGB}{126,47,142}\definecolor{DiffOrange}{RGB}{213,94,0} 
\providecommand{\DIFaddtex}[1]{{\protect\color{DiffPurple}\uwave{#1}}} 
\providecommand{\DIFdeltex}[1]{{\protect\color{DiffOrange}\sout{#1}}} 
\providecommand{\DIFaddbegin}{} 
\providecommand{\DIFaddend}{} 
\providecommand{\DIFdelbegin}{} 
\providecommand{\DIFdelend}{} 
\providecommand{\DIFdelFL}[1]{\DIFdel{#1}} 
\providecommand{\DIFaddbeginFL}{} 
\providecommand{\DIFaddendFL}{} 
\providecommand{\DIFdelbeginFL}{} 
\providecommand{\DIFdelendFL}{} 
\providecommand{\DIFadd}[1]{\texorpdfstring{\DIFaddtex{#1}}{#1}} 
\providecommand{\DIFdel}[1]{\texorpdfstring{\DIFdeltex{#1}}{}} 
\providecommand{\DIFscaledelfig}{0.5}
\RequirePackage{settobox} 
\RequirePackage{letltxmacro} 
\newsavebox{\DIFdelgraphicsbox} 
\newlength{\DIFdelgraphicswidth} 
\newlength{\DIFdelgraphicsheight} 
\LetLtxMacro{\DIFOincludegraphics}{\includegraphics} 
\providecommand{\DIFaddincludegraphics}[2][]{{\color{DiffPurple}\fbox{\DIFOincludegraphics[#1]{#2}}}} 
\providecommand{\DIFdelincludegraphics}[2][]{
\sbox{\DIFdelgraphicsbox}{\DIFOincludegraphics[#1]{#2}}
\settoboxwidth{\DIFdelgraphicswidth}{\DIFdelgraphicsbox} 
\settoboxtotalheight{\DIFdelgraphicsheight}{\DIFdelgraphicsbox} 
\scalebox{\DIFscaledelfig}{
\parbox[b]{\DIFdelgraphicswidth}{\usebox{\DIFdelgraphicsbox}\\[-\baselineskip] \rule{\DIFdelgraphicswidth}{0em}}\llap{\resizebox{\DIFdelgraphicswidth}{\DIFdelgraphicsheight}{
\setlength{\unitlength}{\DIFdelgraphicswidth}
\begin{picture}(1,1)
\thicklines\linethickness{2pt} 
{\color{DiffOrange}\put(0,0){\framebox(1,1){}}}
{\color{DiffOrange}\put(0,0){\line( 1,1){1}}}
{\color{DiffOrange}\put(0,1){\line(1,-1){1}}}
\end{picture}
}\hspace*{3pt}}} 
} 
\LetLtxMacro{\DIFOaddbegin}{\DIFaddbegin} 
\LetLtxMacro{\DIFOaddend}{\DIFaddend} 
\LetLtxMacro{\DIFOdelbegin}{\DIFdelbegin} 
\LetLtxMacro{\DIFOdelend}{\DIFdelend} 
\DeclareRobustCommand{\DIFaddbegin}{\DIFOaddbegin \let\includegraphics\DIFaddincludegraphics} 
\DeclareRobustCommand{\DIFaddend}{\DIFOaddend \let\includegraphics\DIFOincludegraphics} 
\DeclareRobustCommand{\DIFdelbegin}{\DIFOdelbegin \let\includegraphics\DIFdelincludegraphics} 
\DeclareRobustCommand{\DIFdelend}{\DIFOaddend \let\includegraphics\DIFOincludegraphics} 
\LetLtxMacro{\DIFOaddbeginFL}{\DIFaddbeginFL} 
\LetLtxMacro{\DIFOaddendFL}{\DIFaddendFL} 
\LetLtxMacro{\DIFOdelbeginFL}{\DIFdelbeginFL} 
\LetLtxMacro{\DIFOdelendFL}{\DIFdelendFL} 
\DeclareRobustCommand{\DIFaddbeginFL}{\DIFOaddbeginFL \let\includegraphics\DIFaddincludegraphics} 
\DeclareRobustCommand{\DIFaddendFL}{\DIFOaddendFL \let\includegraphics\DIFOincludegraphics} 
\DeclareRobustCommand{\DIFdelbeginFL}{\DIFOdelbeginFL \let\includegraphics\DIFdelincludegraphics} 
\DeclareRobustCommand{\DIFdelendFL}{\DIFOaddendFL \let\includegraphics\DIFOincludegraphics} 
\makeatletter 
\let\sout@orig\sout 
\renewcommand{\sout}[1]{\ifmmode\text{\sout@orig{\ensuremath{#1}}}\else\sout@orig{#1}\fi} 
\makeatother 
\RequirePackage{listings} 
\RequirePackage{color} 
\lstdefinelanguage{DIFcode}{ 
  moredelim=[il][\color{DiffOrange}\sout]{\%DIF\ <\ }, 
  moredelim=[il][\color{DiffPurple}\uwave]{\%DIF\ >\ } 
} 
\lstdefinestyle{DIFverbatimstyle}{ 
	language=DIFcode, 
	basicstyle=\ttfamily, 
	columns=fullflexible, 
	keepspaces=true 
} 
\lstnewenvironment{DIFverbatim}[1][]{\lstset{style=DIFverbatimstyle}}{} 
\lstnewenvironment{DIFverbatim*}[1][]{\lstset{style=DIFverbatimstyle,showspaces=true}}{} 
\begin{document}

\maketitle

\begin{abstract}
Drug-target affinity (DTA) prediction is widely used to prioritize candidate compounds before costly experimental screening. DTA models are often compared under a single data split, even though deployment may require extrapolation to new chemical series, new protein targets, or both. We ask whether the distribution shift used for evaluation changes which architecture appears best. We curate 718,800 unique drug-protein pairs from the ChEMBL and BindingDB datasets. We compare a Morgan-fingerprint + protein-CNN baseline with 12 controlled architectures that combine four drug representations with three ESM-2 interaction modes. Mean validation RMSE increases from 0.950 and 0.945 under scaffold and fingerprint-cluster OOD to 1.299 and 1.321 under protein-cluster and dual OOD. Model rankings are similar across the two chemical shifts ($\tau=0.79$), but agreement with scaffold OOD falls under protein OOD ($\tau=0.39$) and reverses under dual OOD ($\tau=-0.55$). Held-out evaluation, repeated seeds, group-aware bootstrap analysis, and a size-matched control support the same conclusion: architecture selection depends on the form of extrapolation, not only on average error or training-set size. DTA benchmarks should therefore match the chemical and target shifts expected at deployment.
\end{abstract}

\section{Introduction}
Drug-target affinity (DTA) prediction estimates the strength of interaction between a small molecule and a protein target and is widely used to prioritize compounds before costly experimental screening. Early DTA architectures used task-specific sequence and graph encoders (\cite{graphdta2021,moltrans2021}). More recent models combine fixed chemical descriptors or learned molecular encoders with pretrained protein language models and cross-modal interaction mechanisms (\cite{esm22023,dtiam2025,csdta2026,structurefree2025}). These approaches differ in how they represent chemical substructure, protein context, and drug-protein complementarity, making architecture choice an important practical decision.

Drug discovery poses distinct extrapolation problems. In lead optimization and virtual screening, models trained on known binders must score compounds outside the chemical neighborhoods represented in training, including unseen scaffolds and structurally dissimilar molecules. In target identification and drug repurposing, the protein itself may have no close sequence homolog in the labeled training data. These are distinct extrapolation tasks. The first requires transfer of learned structure-activity relationships across chemical space, while the second requires transfer of interaction rules across sequence space. Prior work often describes these as unseen-compound and unseen-target ``cold-start'' settings, in contrast to ``warm-start'' evaluation where both entities are represented during training (\cite{dtiam2025,csdta2026}).

Evaluation protocols can obscure these deployment-relevant differences. Random pair-level splitting can place structurally similar compounds or closely homologous proteins on both sides of the train-test boundary and has been shown to give optimistic estimates of DTA and drug–target interaction (DTI) generalization (\cite{bai2021}). Recent work has likewise emphasized dataset bias, target leakage, and cross-domain performance drops (\cite{debiasdta2023,rigorous2025}). At the same time, the canonical Davis and KIBA benchmarks, which contain approximately 30k and 118k kinase-focused interactions, limit the diversity of the chemical and target space represented in many architecture comparisons (\cite{csdta2026,structurefree2025,davis,kiba}). Larger curated resources such as Papyrus and BindingDB-derived benchmarks broaden this coverage, but make harmonization and split construction central to how generalization is measured (\cite{papyrus2023,drugform2025}). This motivates evaluating architectural designs on a larger, heterogeneous affinity collection under explicitly separated chemical- and target-shift regimes.

Prior work has evaluated chemical- and target-held-out settings, but these studies typically use them to assess a proposed model rather than to ask whether architecture rankings themselves remain stable across evaluation regimes. We address this question using 718,800 drug-protein pairs from ChEMBL and BindingDB (\cite{chembl,bindingdb}). Our main contributions are:

\begin{itemize}
    \item \textbf{To our knowledge, we provide the first systematic DTA evaluation framework centered on architecture-selection stability across complementary scaffold, chemical-similarity, target-sequence, and joint chemical-target distribution shifts.} The four regimes are constructed within a common curated benchmark so that differences in model preference can be attributed to the form of extrapolation being tested.

    \item \textbf{We conduct a controlled factorial comparison of DTA architecture design choices.} We evaluate 12 models spanning four drug representations and three ESM-2 interaction modes, together with a Morgan-fingerprint + protein-CNN baseline. We compare the drug representation and interaction strategy under identical evaluation regimes.

    \item \textbf{We show that the evaluation regime changes model-selection conclusions, not only prediction error.} Architecture rankings remain similar across the two chemical shifts but reorganize under protein and dual OOD, with this pattern supported by held-out evaluation, repeated seeds, Kendall and Spearman rank analyses, group-aware bootstrap uncertainty, split-severity analysis, and a size-matched control.

\end{itemize}

\begin{figure}[t]
\centering
\reviewgraphic[width=\textwidth]{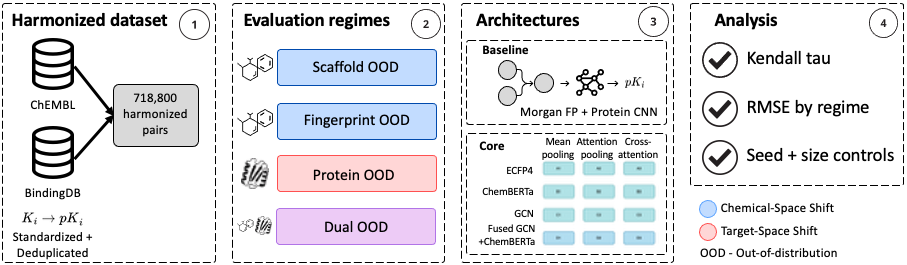}
\caption{Benchmark design. The same architecture family is evaluated under four generalization regimes to independently analyze absolute error and model-selection stability.}
\label{fig:overview}
\end{figure}

\section{Methodology}
\subsection{Dataset}
We combined inhibition constant ($K_i$) measurements from ChEMBL and BindingDB, two large public databases of experimentally measured small-molecule--protein bioactivity and binding-affinity data, and converted them to $pK_i=-\log_{10}(K_i[\mathrm{M}])$. Rows with invalid molecular structures, missing protein sequences, or missing affinity values were removed. Simplified Molecular Input Line Entry System (SMILES) strings were standardized with RDKit, a toolkit used to analyze and manipulate chemical data. Repeated measurements for the same drug-protein pair were then identified and mean-aggregated before split construction.

The final dataset contains \textbf{718,800 unique drug-protein pairs}, spanning \textbf{273,528 compounds} and \textbf{3,704 proteins}. Median $pK_i$ is 6.48 (mean 6.61, SD 1.52), with 616,359 pairs (85.7\%) satisfying $pK_i>5$ ($K_i<10\mu \mathrm{M}$). 

\subsection{Evaluation regimes}
We evaluate model generalization under four increasingly challenging out-of-distribution (OOD) regimes that hold out chemical structure, chemical similarity, protein sequence similarity, or both drug and protein distributions.

\textbf{Scaffold OOD.} Bemis-Murcko scaffolds are computed using RDKit and assigned group-wise to train, validation, or test partitions.

\textbf{Fingerprint-cluster OOD.} Compounds are represented using 2048-bit extended-connectivity fingerprints with a maximum diameter of 4 bonds (ECFP4) \DIFdelbegin \DIFdel{. }\DIFdelend [Bai et al., 2021; Simm et al., 2021]\DIFaddbegin \DIFadd{. }\DIFaddend Sequential sphere-exclusion clustering uses a Tanimoto similarity threshold of 0.60, with each compound joining the first qualifying center or seeding a new cluster. Entire clusters are assigned to one partition.

\textbf{Protein-cluster OOD.} Protein sequences are represented by amino-acid 3-mer sets and clustered using Jaccard similarity at threshold 0.40. Each protein cluster is assigned entirely to one partition.

\textbf{Dual OOD.} Chemical groups and protein clusters are first assigned independently to train, validation, or test. We retain an interaction only when both assignments agree. A test interaction therefore contains a drug from a held-out chemical group and a protein from a held-out sequence cluster, so neither side of the pair is represented in training. This directly tests simultaneous extrapolation across both axes. Conflicting assignments are removed, which reduces the retained training set. Section~\ref{sec:sizecontrol} tests whether that reduction alone explains the ranking change.

\subsection{Architecture matrix and training}
We first evaluate a Morgan-fingerprint + character-level protein Convolutional Neural Network (CNN) baseline under all four split definitions. This provides a non-pretrained protein encoder outside the Evolutionary Scale Modeling-2 (ESM-2) design space. The controlled matrix then contains 12 architectures formed from four drug representations and three protein interaction modes. The drug representations are ECFP4, ChemBERTa, GCN, and fused GCN+ChemBERTa. All three protein modes use ESM-2 and differ in aggregation through mean pooling, learned attention pooling, or drug-conditioned cross-attention. This $4\times3$ design separates drug representation choice from interaction strategy under a common evaluation protocol. 

Architectures are trained independently within each regime and ranked by RMSE; ranking agreement is measured primarily with Kendall's $\tau$. Spearman's rank correlation ($\rho$) is additionally used as a secondary validation of rank stability across evaluation regimes. Split-severity statistics, optimization settings, sample-size controls, rank-correlation analyses, and architecture-specific training details are reported in Appendix~\ref{app:study_details}.
\section{Results}
\subsection{Distribution shift dominates absolute error}
Across the 12 architectures, performance is similar under the two drug-based distribution shifts. Mean validation RMSE is slightly lower under fingerprint OOD than scaffold OOD (0.945 vs. 0.950) despite fingerprint OOD having lower chemical similarity to the training set. Error increases substantially when generalizing to unseen proteins: mean RMSE rises to 1.299 under protein OOD and 1.321 when both drugs and proteins are OOD, corresponding to increases of 36.7\% and 39.0\% relative to scaffold OOD. Detailed architecture-level results are provided in Appendix~\ref{app:additional_results}.

These results identify target-sequence shift as the dominant difficulty increase in the current benchmark. Dual OOD is hardest overall, but only modestly exceeds protein OOD despite adding substantial chemical separation. \DIFdelbegin 

\DIFdelend Held-out evaluation of all 12 architectures preserves this severity pattern, with a mean RMSE of 0.986, 0.946, 1.354, and 1.368 under scaffold, fingerprint, protein, and dual OOD, respectively (Table~\ref{tab:testmatrix}). The split-specific winner is unchanged for scaffold, fingerprint, and dual OOD, while protein OOD changes from ECFP+AttnPool on validation to GCN+ChemBERTa+CrossAttn on the held-out test set.

\subsection{Architecture rankings depend on the regime}
Scaffold and fingerprint OOD produce similar model orderings ($\tau=0.79$, 7 of 66 pairwise reversals). Agreement falls between scaffold and protein OOD ($\tau=0.39$, 20 reversals) and becomes negative between scaffold and dual OOD ($\tau=-0.55$, 51 reversals). Fingerprint versus dual OOD shows the largest inversion ($\tau=-0.64$, 54 reversals). The top three architectures under scaffold and dual OOD do not overlap. Spearman rank correlations independently reproduce this pattern, with strong agreement between scaffold and fingerprint OOD ($\rho=0.90$) and negative correlations between the chemical and dual-OOD rankings ($\rho=-0.73$ for scaffold and $-0.77$ for fingerprint). Full pairwise results are provided in Appendix~\ref{app:rankcorrelation}.

\begin{figure}[t]
\centering
\reviewgraphic[width=0.85\textwidth]{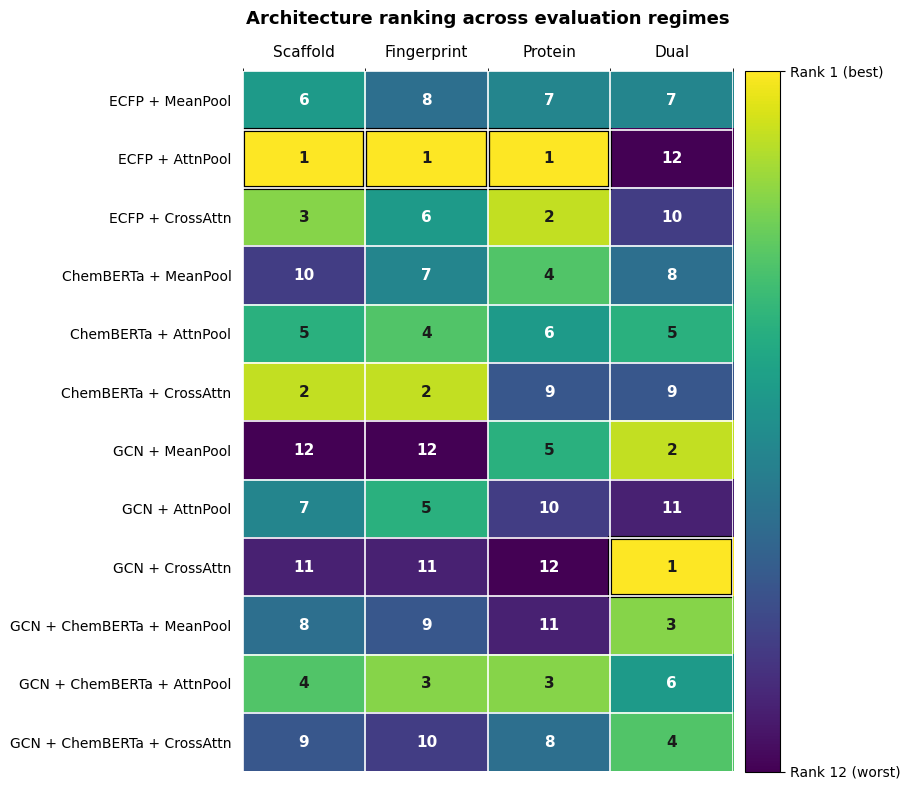}
\caption{Architecture-rank trajectories under RMSE. Architecture rankings are similar under scaffold and fingerprint OOD but shift substantially under protein and dual OOD. This figure visualizes the paper's central model-selection result.}
\label{fig:ranks}
\end{figure}

The clearest inversion is between ECFP+AttnPool and GCN+CrossAttn. ECFP+AttnPool ranks first under scaffold, fingerprint, and protein OOD but last under dual OOD. GCN+CrossAttn ranks near the bottom in the first three regimes and first under dual OOD. Their relative advantage changes from 0.104 RMSE in favor of ECFP+AttnPool under scaffold OOD to 0.092 in favor of GCN+CrossAttn under dual OOD (Figure \ref{fig:ranks}).

Repeated-seed results, the Morgan-fingerprint + protein-CNN comparison, group-aware bootstrap uncertainty, and the training-size ablation are moved to Appendix~\ref{app:robustness}.

\section{Discussion and conclusion}
Target-sequence shift produces the largest increase in prediction error in this benchmark. Moving from chemical OOD to protein OOD increases average RMSE substantially more than changing between scaffold- and fingerprint-based chemical shifts. Architecture preference also changes across regimes. The Morgan-fingerprint + protein-CNN baseline shows the same regime dependence outside the controlled ESM-2 matrix, ranking lowest on the chemical-OOD validation splits, highest under protein OOD, and lowest again under dual OOD.

The strongest reversal occurs under dual OOD. Repeated initialization seeds preserve its direction. Five size-matched protein-OOD subsets do not reproduce the GCN advantage, while group-aware bootstrap intervals favor GCN under dual OOD. Together with the split-severity audit, these controls support an architecture-specific response to simultaneous chemical and target-sequence shift rather than a simple training-size effect. This pattern is consistent with different architectural inductive biases becoming useful under different forms of extrapolation.

These findings also suggest a potential mitigation through similarity-aware model selection, where nearest-training chemical similarity (ECFP4 Tanimoto) and target-sequence similarity (3-mer Jaccard) are estimated before prediction and used to select among architectures with different OOD strengths. In our benchmark, dual OOD is characterized by low similarity on both axes (median Tanimoto 0.619; median Jaccard 0.142), whereas protein OOD retains chemically familiar compounds (median Tanimoto 1.000) while shifting target sequences (median Jaccard 0.139), suggesting that these similarity scores could serve as routing signals for selecting models suited to the expected form of extrapolation.

This distinction matters for model selection in drug discovery. A benchmark that preserves familiar compounds but holds out target neighborhoods asks a different scientific question from one that simultaneously moves into new chemical and target space. Reporting a single aggregate benchmark number can therefore obscure which inductive biases are favored by the intended deployment setting. Our results motivate evaluating DTA models under explicitly characterized chemical and target shifts and reporting ranking stability, not only absolute error.

The study remains retrospective and sequence-based. Public ChEMBL and BindingDB measurements combine heterogeneous assay contexts, and the present split definitions remain proxies for prospective discovery. We plan to test whether the ranking behavior extends to larger pretrained encoders and alternative graph architectures.

\newpage
\bibliographystyle{plainnat}
\setlength{\bibsep}{2pt plus 3pt}
\IfFileExists{references.bib}{\bibliography{references}}{%

\begin{thebibliography}{1}
  \bibitem{missingreferences}Bibliography file \texttt{references.bib} was not supplied with the compared source files.
  \end{thebibliography}
  
}

\appendix

\section{Detailed study design}
\label{app:study_details}

\subsection{Split-severity audit}
Nearest-training similarity is measured after partitioning. On the chemical axis, median nearest-training ECFP4 Tanimoto is 0.719 for scaffold OOD, 0.676 for fingerprint OOD, 1.000 for protein OOD, and 0.619 for dual OOD. Thus, the protein split leaves the compound side familiar, while dual OOD adds a substantial chemical-space shift. On the target axis, scaffold and fingerprint splits have a median nearest-training 3-mer Jaccard of 1.00, whereas protein and dual OOD reduce this value to 0.139 and 0.142. Approximately 32\% of test proteins in both protein-focused regimes have nearest-training Jaccard below 0.10.

\begin{figure}[H]
\centering
\reviewgraphic[width=0.78\textwidth]{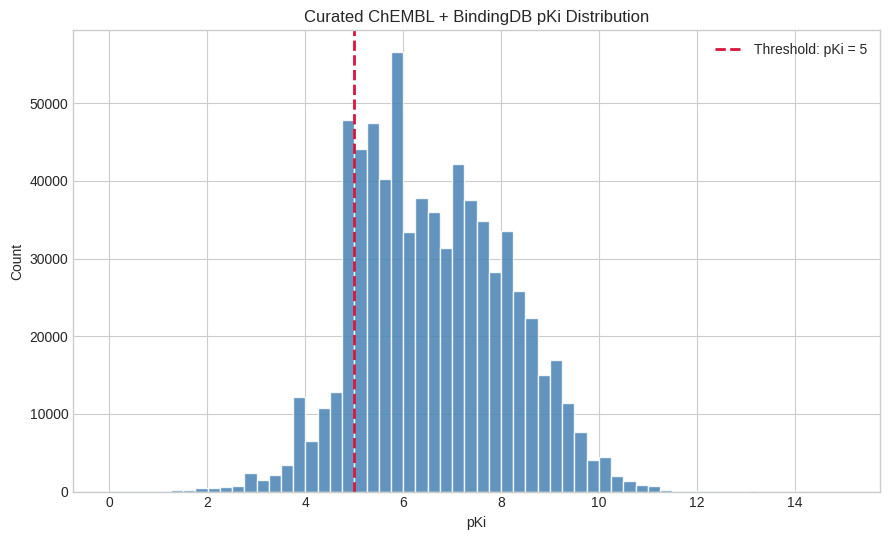}
\caption{Distribution of $pK_i$ values in the curated ChEMBL and BindingDB dataset. A reference line at $pK_i=5$ may be included to show the corresponding $K_i=10\,\mu\mathrm{M}$ location without defining it as a universal activity threshold.}
\label{fig:pki_distribution}
\end{figure}

\subsection{Training and analysis details}
Architectures are trained independently within each regime and ranked by RMSE. Rank agreement between split pairs is quantified with Kendall's $\tau$. With 12 architectures, each ranking contains 66 architecture pairs. Three representative architectures are retrained across three random seeds. The training-size control downsamples the protein-OOD training partition from approximately 511k to 352k pairs to match the dual-OOD training size.

All models are optimized with AdamW using weight decay $1\times10^{-4}$ and cosine annealing, with a maximum of 40 epochs and early-stopping patience of 7 epochs based on RMSE. Learning rates are $1\times10^{-3}$ for ECFP/GCN branches and $5\times10^{-4}$ for branches containing ChemBERTa. ESM-2 (\texttt{esm2\_t12\_35M\_UR50D}, 480 dimensions) and ChemBERTa (\texttt{DeepChem/ChemBERTa-77M-MTR}, 384 dimensions) are frozen during training. Only task-specific graph, projection, interaction, and regression modules are optimized. Architecture-specific batch sizes range from 64 to 2048, and trainable parameter counts range from 0.62M to 2.09M (Table~\ref{tab:hyperparams}).

\subsection{Rank-correlation analysis}
\label{app:rankcorrelation}

As a secondary check of ranking stability, we computed Spearman's rank correlation ($\rho$) between the architecture rankings induced by each pair of evaluation regimes. The results reproduce the pattern observed with Kendall's $\tau$. Scaffold and fingerprint OOD show strong agreement ($\rho=0.895$), whereas agreement weakens when either chemical regime is compared with protein OOD ($\rho=0.448$ and $0.434$). Rankings become negatively correlated under dual OOD, with $\rho=-0.727$ relative to scaffold OOD and $\rho=-0.769$ relative to fingerprint OOD. Protein and dual OOD are also negatively correlated ($\rho=-0.476$). Thus, the observed ranking reorganization is not specific to the choice of Kendall's $\tau$ as the primary rank-agreement statistic.

\begin{table}[H]
\centering
\caption{Spearman rank correlations ($\rho$) between architecture rankings across evaluation regimes. Positive values indicate similar architecture orderings and negative values indicate inverted orderings.}
\label{tab:spearman}
\small
\setlength{\tabcolsep}{7pt}
\begin{tabular}{lccc}
\toprule
Regime pair & Spearman $\rho$ \\
\midrule
Scaffold vs. Fingerprint & 0.895 \\
Scaffold vs. Protein & 0.448 \\
Scaffold vs. Dual OOD & -0.727 \\
Fingerprint vs. Protein & 0.434 \\
Fingerprint vs. Dual OOD & -0.769 \\
Protein vs. Dual OOD & -0.476 \\
\bottomrule
\end{tabular}
\end{table}

\section{Robustness and ablation analyses}
\label{app:robustness}

\subsection{Additional CNN baseline}
A Morgan-fingerprint + protein-CNN baseline shows the same regime-dependent behavior beyond the ESM-2 factorial grid. Its validation RMSE increases from 1.0355/1.0968 under scaffold/fingerprint OOD to 1.2505 under protein OOD and 1.4700 under dual OOD. It underperforms all 12 core architectures under scaffold and fingerprint OOD, ranks first under protein OOD, and last under dual OOD.

\begin{table}[H]
\centering
\caption{Morgan-fingerprint + protein-CNN baseline. Values are RMSE under the same split definitions and training protocol as the core matrix.}
\label{tab:cnnbase}
\small
\setlength{\tabcolsep}{6pt}
\begin{tabular}{lcc}
\toprule
Split & Validation RMSE & Held-out test RMSE \\
\midrule
Scaffold & 1.0355 & 1.1175 \\
Fingerprint & 1.0968 & 1.1802 \\
Protein & \textbf{1.2505} & 1.3359 \\
Dual OOD & 1.4700 & 1.4853 \\
\bottomrule
\end{tabular}
\end{table}

\subsection{Repeated-seed robustness}
Three representative architectures were repeated over three seeds. ECFP+AttnPool achieves RMSE of $0.8810\pm0.0022$, $1.2544\pm0.0145$, and $1.3652\pm0.0358$ under scaffold, protein, and dual OOD, respectively. GCN+CrossAttn obtains $0.9947\pm0.0076$, $1.3297\pm0.0272$, and $1.3203\pm0.0385$. Thus ECFP+AttnPool remains better under scaffold and protein OOD, whereas GCN+CrossAttn is better on average under dual OOD. ChemBERTa+AttnPool provides an intermediate reference at $0.9305\pm0.0041$, $1.3053\pm0.0105$, and $1.3242\pm0.0092$.

\subsection{Training-size ablation}
\label{sec:sizecontrol}
Dual OOD retains approximately 352k training pairs compared with about 511k under protein OOD. To test whether this difference explains the ranking reversal, we repeatedly downsampled the protein-OOD training set to 352k pairs. Across five independent draws, ECFP+AttnPool and GCN+CrossAttn perform similarly under protein OOD ($1.2804\pm0.0246$ vs. $1.2917\pm0.0315$). Conversely, under dual OOD at the same 352k training size, GCN+CrossAttn outperforms ECFP+AttnPool by 0.0924 RMSE (Table~\ref{tab:sizecontrol}). Reduced training size alone therefore does not reproduce the dual-OOD reversal.

\begin{table}[H]
\centering
\caption{Training-size ablation for ECFP+AttnPool and GCN+CrossAttn. Protein-matched values are mean $\pm$ SD across five independent 352k training subsets. Values are RMSE.}
\label{tab:sizecontrol}
\small
\setlength{\tabcolsep}{5pt}
\begin{tabular}{lccc}
\toprule
Model & Protein 511k & Protein 352k & Dual 352k \\
\midrule
ECFP+AttnPool & \textbf{1.2728} & $\mathbf{1.2804\pm0.0246}$ & 1.3652 \\
GCN+CrossAttn & 1.3595 & $1.2917\pm0.0315$ & \textbf{1.2728} \\
\bottomrule
\end{tabular}
\end{table}

\subsection{Group-aware bootstrap uncertainty}
We quantified the ECFP+AttnPool versus GCN+CrossAttn difference with 10,000 paired bootstrap resamples at the corresponding split-group level. Negative RMSE differences favor ECFP and positive differences favor GCN. The 95\% CIs favor ECFP under scaffold OOD ($-0.1113$, $[-0.1275,-0.0956]$) and fingerprint OOD ($-0.0860$, $[-0.1014,-0.0710]$). The protein-OOD interval includes zero ($-0.0555$, $[-0.2306,0.0968]$). Under dual OOD, the sign reverses and the interval excludes zero ($+0.2059$, $[0.0153,0.3851]$), favoring GCN+CrossAttn.

\section{Architecture and training details}
\label{app:architecture}

\subsection{Model configuration}

\paragraph{Frozen pretrained representations.} ESM-2 uses \texttt{esm2\_t12\_35M\_UR50D} (480 dimensions), and
ChemBERTa uses \texttt{DeepChem/ChemBERTa-77M-MTR} (384 dimensions).
Both pretrained encoders are frozen during training so that the comparison
focuses on drug representation and interaction design rather than
differential end-to-end fine-tuning.

\begin{table}[th]
\centering
\caption{Architecture-specific batch sizes and trainable parameter counts. Pretrained ESM-2 and ChemBERTa representations are frozen. Parameter counts refer only to trainable task-specific modules.}

\label{tab:hyperparams}
\small
\setlength{\tabcolsep}{6pt}
\begin{tabular}{llllr}
\toprule
Architecture & Drug encoder & Protein / fusion mode & Batch size & Trainable params \\
\midrule
ECFP+MeanPool & ECFP4 & MeanPool MLP & 2048 & 1.04M \\
ChemBERTa+MeanPool & ChemBERTa & MeanPool MLP & 2048 & 0.62M \\
ECFP+AttnPool & ECFP4 & Dot-product AttnPool & 128 & 1.17M \\
ECFP+CrossAttn & ECFP4 & CrossAttn & 128 & 2.09M \\
ChemBERTa+AttnPool & ChemBERTa & Dot-product AttnPool & 128 & 0.74M \\
ChemBERTa+CrossAttn & ChemBERTa & CrossAttn & 128 & 1.66M \\
GCN+MeanPool & GCN & MeanPool MLP & 256 & 0.72M \\
GCN+AttnPool & GCN & Dot-product AttnPool & 64 & 0.84M \\
GCN+CrossAttn & GCN & CrossAttn & 64 & 1.77M \\
GCN+ChemBERTa+MeanPool & GCN+ChemBERTa & MeanPool MLP & 128 & 0.77M \\
GCN+ChemBERTa+AttnPool & GCN+ChemBERTa & Dot-product AttnPool & 64 & 0.89M \\
GCN+ChemBERTa+CrossAttn & GCN+ChemBERTa & CrossAttn & 64 & 1.82M \\
\bottomrule
\end{tabular}
\end{table}

\section{Additional results}
\label{app:additional_results}

\subsection{Validation RMSE across architectures}

\begin{table}[H]
\centering
\caption{Validation RMSE across the architecture-by-split
matrix. Lower is better. Bold indicates the best model within each regime;
the final row gives the column mean across all 12 architectures.}
\label{tab:matrix}
\small
\setlength{\tabcolsep}{6pt}
\begin{tabular}{lcccc}
\toprule
Architecture & Scaffold & Fingerprint & Protein & Dual OOD \\
\midrule
ECFP + MeanPool & 0.9523 & 0.9511 & 1.2977 & 1.3221 \\
ChemBERTa + MeanPool & 0.9633 & 0.9490 & 1.2858 & 1.3286 \\
ECFP + AttnPool & \textbf{0.8815} & \textbf{0.9034} & \textbf{1.2728} & 1.3652 \\
ECFP + CrossAttn & 0.9286 & 0.9441 & 1.2749 & 1.3604 \\
ChemBERTa + AttnPool & 0.9362 & 0.9250 & 1.2925 & 1.3112 \\
ChemBERTa + CrossAttn & 0.9192 & 0.9186 & 1.3091 & 1.3341 \\
GCN + MeanPool & 1.0265 & 0.9893 & 1.2904 & 1.2837 \\
GCN + AttnPool & 0.9533 & 0.9369 & 1.3105 & 1.3623 \\
GCN + CrossAttn & 0.9851 & 0.9754 & 1.3595 & \textbf{1.2728} \\
GCN+ChemBERTa + MeanPool & 0.9603 & 0.9564 & 1.3150 & 1.2920 \\
GCN+ChemBERTa + AttnPool & 0.9332 & 0.9239 & 1.2808 & 1.3168 \\
GCN+ChemBERTa + CrossAttn & 0.9624 & 0.9632 & 1.3004 & 1.3009 \\
\midrule
Mean & 0.9502 & 0.9447 & 1.2991 & 1.3212 \\
\bottomrule
\end{tabular}
\end{table}

\subsection{Held-out test RMSE across architectures}
\begin{table}[H]
\centering
\caption{Held-out test RMSE across the architecture-by-split matrix. Lower is better. Bold indicates the best model within each regime; the final row gives the column mean across all 12 architectures.}
\label{tab:testmatrix}
\small
\setlength{\tabcolsep}{6pt}
\begin{tabular}{lcccc}
\toprule
Architecture & Scaffold & Fingerprint & Protein & Dual OOD \\
\midrule
ECFP + MeanPool & 0.9872 & 0.9615 & 1.3823 & 1.3599 \\
ECFP + AttnPool & \textbf{0.9234} & \textbf{0.8905} & 1.3713 & 1.4948 \\
ECFP + CrossAttn & 0.9738 & 0.9551 & 1.3703 & 1.4349 \\
ChemBERTa + MeanPool & 0.9969 & 0.9469 & 1.3360 & 1.4143 \\
ChemBERTa + AttnPool & 0.9622 & 0.9307 & 1.4080 & 1.3625 \\
ChemBERTa + CrossAttn & 0.9594 & 0.9228 & 1.3098 & 1.3883 \\
GCN + MeanPool & 1.0660 & 0.9850 & 1.2864 & 1.3052 \\
GCN + AttnPool & 0.9897 & 0.9308 & 1.4126 & 1.3699 \\
GCN + CrossAttn & 1.0347 & 0.9766 & 1.4268 & \textbf{1.2889} \\
GCN+ChemBERTa + MeanPool & 0.9806 & 0.9610 & 1.3655 & 1.3330 \\
GCN+ChemBERTa + AttnPool & 0.9637 & 0.9223 & 1.3508 & 1.3513 \\
GCN+ChemBERTa + CrossAttn & 0.9924 & 0.9693 & \textbf{1.2263} & 1.3085 \\
\midrule
Mean & 0.9858 & 0.9460 & 1.3538 & 1.3676 \\
\bottomrule
\end{tabular}
\end{table}

\end{document}